# A Lightweight Dual-Branch System for Weakly-Supervised Video Anomaly Detection on Consumer Edge Devices

Wen-Dong Jiang, *Graduate student Member*, *IEEE*, Chih-Yung Chang, *Member, IEEE*, Ssu-Chi Kuai, and Diptendu Sinha Roy, *Senior Member, IEEE*

*Abstract*—**The growing demand for intelligent security in consumer electronics, such as smart home cameras and personal monitoring systems, is often hindered by the high computational cost and large model sizes of advanced AI. These limitations prevent the effective deployment of real-time Video Anomaly Detection (VAD) on resource-constrained edge devices. To bridge this gap, this paper introduces Rule-based Video Anomaly Detection (RuleVAD), a novel, lightweight system engineered for high-efficiency and low-complexity threat detection directly on consumer hardware. RuleVAD features an innovative decoupled dual-branch architecture to minimize computational load. An implicit branch uses visual features for rapid, coarse-grained binary classification, efficiently filtering out normal activity to avoid unnecessary processing. For potentially anomalous or complex events, a multimodal explicit branch takes over. This branch leverages YOLO-World to detect objects and applies data mining to generate interpretable, text-based association rules from the scene. By aligning these rules with visual data, RuleVAD achieves a more nuanced, fine-grained classification, significantly reducing the false alarms common in vision-only systems. Extensive experiments on the XD-Violence and UCF-Crime benchmark datasets show that RuleVAD achieves superior performance, surpassing existing state-of-the-art methods in both accuracy and speed. Crucially, the entire system is optimized for low-power operation and is fully deployable on an NVIDIA Jetson Nano board, demonstrating its practical feasibility for bringing advanced, real-time security monitoring to everyday consumer electronic devices.**

## I. INTRODUCTION

Video anomaly detection aims to promptly detect abnormal behaviors in surveillance videos and issue warnings to maintain public safety, serving as a crucial component in the realization of smart cities [1]. With the rapid advancement of artificial intelligence, researchers are actively exploring the use of deep learning technologies in surveillance systems to overcome the inefficiencies and lack of robustness in traditional manual monitoring [2][3].

In recent years, Weakly-Supervised Video Anomaly Detection (WVAD) has emerged as a significant research area. Unlike traditional methods that require frame-by-frame labeling and supervised learning for training, WVAD can achieve frame-level anomaly monitoring based solely on video-level annotations [4][5].

However, relying solely on visual features often encounters bottlenecks. Consider a common scenario: in a crowded square, an individual excitedly raises their arms and shouts. A vision-only model might associate this motion pattern with violent behaviors like "fighting" or "quarreling" from its database, thus triggering a false alarm. In reality, this person could simply be cheering for a distant team's goal or holding a sign that reads "Long Live Peace." In such cases, the true intent of the behavior (e.g., "cheering" or "peaceful demonstration") is embedded in semantic information beyond the visual data. This highlights the importance of introducing multimodal models with text comprehension capabilities.

Current mainstream methods typically use pre-trained models such as Inflated 3D ConvNet (I3D) [6], Transformer [7], and Contrastive Language-Image Pre-training (CLIP) [8] to extract features from full-resolution frames. These features are then processed by temporal models, and finally, a Multiple Instance Learning (MIL) mechanism is applied to train a classifier for frame-level anomaly prediction [9] [10] [11]. However, the system design of these methods generally adopts a highly-coupled, end-to-end paradigm. This approach tends to generate substantial computational redundancy in practice, significantly increasing the system's overall complexity and raising concerns about its application in real-world scenarios [12]-[16].

To address these issues, this paper proposes a system named RuleVAD. As shown in Fig. 1, unlike previous studies that directly adopt coupled end-to-end designs, our proposed RuleVAD employs a differentiated, decoupled design for image and text modalities. It introduces a dual-branch architecture to achieve an end-to-end violence surveillance system with lower computational complexity. The first branch, called the implicit branch, focuses on coarse-grained binary classification using visual features. Within this branch, image feature extraction is divided into two channels: one for extracting scene frames and another specifically for behavior extraction. The second branch, known as the explicit branch, performs fine-grained classification by leveraging the alignment between text and images. For the text channel in the explicit branch, RuleVAD

This work was supported by the National Science and Technology Council of Taiwan, Republic of China, under grant number NSTC 112-2622-E-032-005. (Corresponding author: Chih-Yung Chang.)

Wen-Dong Jiang is currently pursuing a Ph.D. degree with the Department of Computer Science and Information Engineering, Tamkang University, New Taipei 25137, Taiwan. (e-mail: wendongjiang@ieee.org).

Chih-Yung Chang is with the Department of Computer Science and Information Engineering, Tamkang University, New Taipei 25137, Taiwan. (email: cychang@mail.tku.edu.tw).

Ssu-Chi Kuai is with the Department of Information Management, National Taipei University of Business, Taipei 100025, Taiwan. (email: sckuai@ntub.edu.tw).

Diptendu Sinha Roy is with the Department of Computer Science and Engineering, National Institute of Technology, Shillong, 793003, India (e-mail: diptendu.sr@nitm.ac.in).

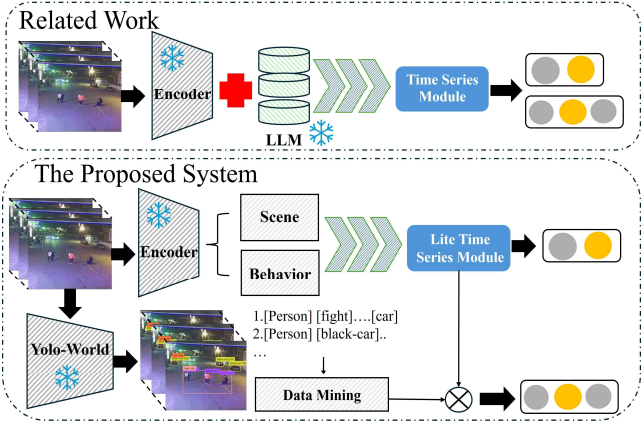

**Fig. 1.** Differences from previous studies.

integrates YOLO-World [20] to detect objects in video frames and applies data mining techniques to identify association rules. These rules provide interpretable, rule-based descriptions of the video content, thereby assisting the model in performing more detailed classification. Furthermore, RuleVAD incorporates a lightweight time-series module to support system deployment in edge computing scenarios. To validate the effectiveness of our system, we conducted experiments on two benchmark datasets: XD-Violence and UCF-Crime.

In summary, the contributions of this paper can be summarized into the following three points:

1. **Proposed an innovative Decoupled Dual-Branch Architecture.** Unlike traditional, highly-coupled models, this system separates visual violence detection (the implicit branch) from textual semantic understanding (the explicit branch). This design significantly reduces computational complexity and resource consumption, addressing the efficiency bottlenecks of existing methods in practical applications.

2. **Introduced a Rule-Based Semantic Understanding method.** By combining YOLO-World object detection with data mining techniques, the system automatically generates interpretable "rules" to understand a scene (e.g., the co-occurrence of a "person" and a "peace sign"). This effectively resolves ambiguities that arise from relying solely on visual features, improving judgment accuracy in complex scenarios.

3. **Focused on a Lightweight Design for Edge Computing.** The system integrates a lightweight time-series module, specifically optimized for deployment on resource-constrained edge devices. This gives the high-accuracy anomaly detection system the practicality and feasibility for real-world deployment.

The remainder of the paper is organized as follows. Section II discusses and compares previous relevant studies. Section III describes the assumptions and problem formulation in detail. Section IV details the proposed interpretable multimodal model. Section V provides experiments and performance evaluation. The conclusions are discussed in Section VI.

## II. RELATED WORK

This chapter presents a review of some relevant studies.

These studies are categorized into two parts: Weakly Supervised Video Anomaly Detection and Visual Language Pre-Training Models.

### A. Weakly Supervised Video Anomaly Detection

*Sultani et al.* [6] and *Hasan et al.* [22] were the first to introduce the MIL model for violence monitoring in supervised video anomaly detection. The proposed approaches packaged surveillance videos into a package and then input them into I3D and C3D networks for binary classification of violent and non-violent events. *Ji and Lee* [23] proposed the One-Class Support Vector Machine (OCSVM) model for video anomaly detection. However, due to the limitations of 3D network-based designs in terms of real-time performance and hardware requirements in practical applications, research gradually shifted towards combining video frame extraction and time series modules for violence monitoring. Subsequent research began to focus on using self-attention mechanisms, Transformer, or Graph ConvNet (GCN) to capture temporal and contextual relationships in video content. *Zhong et al.* [24] proposed a GCN-based method to calculate feature similarity and temporal consistency between monitored video segments for the monitoring of violent behavior. While *Tian et al.* [25] utilized a self-attention network to capture the global temporal context of violent behavior in monitoring videos. Their proposed model, referred to as Robust Temporal Feature Magnitude (RTFM), effectively leverages temporal information to enhance the monitor of violent actions. *Wu et al.* [4] proposed Hierarchical Learning Network (HL-NET), which developed a global and local attention module to identify temporal dependencies and thereby obtain more expressive embeddings from monitoring videos.

Those approaches for Weakly Supervised Video Anomaly Detection typically employed a coupled, end-to-end architecture that detected anomalies by predicting the probability of abnormal frames. A primary drawback was that these systems featured highly coupled designs without computational optimization, which led to extremely high computational complexity. Moreover, their time-series modules predominantly used similarity calculations based on vector flattening. This methodology introduced significant computational redundancy and made the approaches ill-suited for lightweight computing scenarios.

### B. Visual language pre-training Models

In some practical application scenarios, only single-modal cameras were typically available, and traditional multi-modal designs were often not applicable to these single-modal scenarios. Thanks to certain pre-trained models, such as CLIP [8], which aligned text and images to accomplish multi-modal tasks, it became possible to achieve multi-modal tasks even in scenarios where only a single modality was present. For instance, *Zhou et al.* [26] introduced an enhanced image classification approach leveraging the CLIP, which significantly improved performance in complex visual recognition tasks. *Mokady et al.* [27] made a major advancement in automatic image captioning by refining the integration of vision and language models. *Zhou et al.* [28]

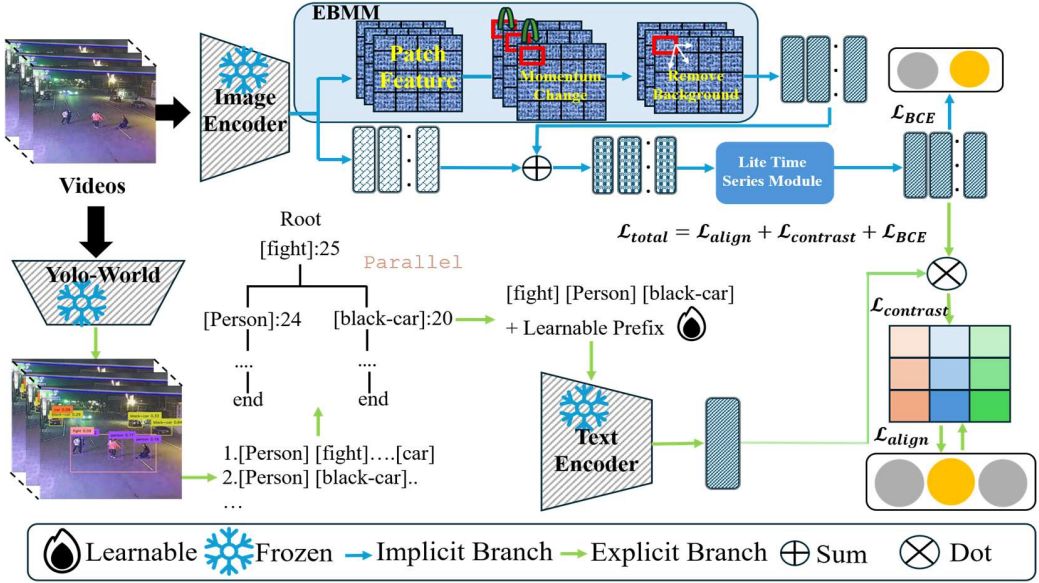

**Fig. 2.** The pipeline of the proposed RuleVAD system. The Implicit Branch (top) performs rapid, coarse-grained binary classification by processing visual features through parallel behavior and scene channels, followed by a lightweight temporal module to filter normal events. The Explicit Branch (bottom) conducts fine-grained, multi-class classification by aligning video features (from the implicit branch) with text-based rules generated via YOLO-World and data mining, using contrastive learning to achieve nuanced understanding.

applied CLIP to object monitoring, improving the model's ability to distinguish between similar objects. *Yu et al.* [29] made notable progress in scene text monitoring, particularly in enhancing the accuracy and efficiency of text localization in cluttered environments. Additionally, *Rao et al.* [30] conducted valuable research on dense prediction tasks, focusing on improving the precision of pixel-level predictions in dense visual data. *Lv et al.* [31] proposed Uncertainty-aware Multiple Instance Learning (UMIL), which incorporated uncertainty estimation into the multiple instances learning framework to improve the system's robustness and accuracy. *Joo et al.* [32] proposed CLIP with Temporal Shift Attention (CLIP-Tsa), which introduces a temporal shift attention mechanism to better capture temporal dependencies in video data, enhancing the model's performance in tasks. In Anomaly Detection, *Luo et al.* [33] proposed CLIP4Clip, a method aimed at transferring the knowledge of CLIP models to the field of video-text retrieval; *Wu et al.* [18] proposed Video Anomaly Monitor CLIP (VadCLIP), a simple yet powerful baseline that efficiently adapts pre-trained image-based visual-language models to leverage their robust capabilities in general video understanding. *Wu et al.* [17] introduced Spatio-Temporal Prompting (STPrompt), a CLIP-based three-branch architecture to address classification and localization in video anomaly detection.

Although CLIP-based pretrained models achieved excellent performance in image-text matching tasks, their design often leaned more heavily towards the image branch. The text branch, in contrast, relied on a LLM to generate tokens, which were then converted into embeddings via an encoder. The challenges of this design became particularly apparent in video tasks. Video descriptions were often represented as overly abstract, high-dimensional embeddings. Even when tokens were generated, they could not be easily formed into comprehensible natural language descriptions; instead, they typically represented highly compressed information. Furthermore, the method of aggregating frame-level descriptions into a video-level representation via simple similarity calculations introduced excessive computational redundancy. Moreover, the output from the LLM could be inconsistent and unstable in certain cases, which further compromised the accuracy of the text branch.

## III. The Proposed RuleVAD

This section presents the proposed RuleVAD, a system designed for low-cost, weakly supervised video anomaly detection. Unlike previous works [4][26][31][32] that designed coupled end-to-end systems by integrating multiple pre-trained models, RuleVAD adopts a decoupled strategy. It features implicit and explicit branches and employs a multi-channel approach to achieve interpretable video anomaly detection. This design allows each branch to handle its respective task independently during training, free from interference and thereby improving operational speed.

Furthermore, regarding the textual branch, our approach diverges from previous methods [17],[18],[31],[32] that used LLMs to generate frame-by-frame descriptions before aggregation with a temporal module. Instead, RuleVAD combines a pre-trained object detection model with data mining techniques to derive descriptive rules for videos. This enables RuleVAD to maintain strong interpretability while simultaneously reducing computational redundancy during both training and deployment. Finally, compared to other systems [4],[6],[17],[18],[22],[25], RuleVAD utilizes a lightweight temporal module, allowing the model to consume significantly fewer computational resources throughout its training and deployment phases."

Figure 2 illustrates the detailed architecture of the proposed RuleVAD system. The system consists of two branches: the Implicit Branch and the Explicit Branch. Each branch has two respective channels. The Implicit Branch utilizes visual features and a lightweight temporal sequence module to perform coarse-grained binary classification. It is designed to jointly extract spatial information at the frame level through both the behavior and scene channels, then analyze the temporal relationships using the lightweight temporal module and finally train the branch through a binary classifier. Then the explicit Branch achieves fine-grained multi-class classification using both text and images, with its two channels corresponding to text and video, respectively. The text is derived from YOLO-World and association rules generated by data mining methods, while the video is represented by the embedding produced by the Implicit Branch before executing binary classification. The text and video embeddings are combined through contrastive learning in high-dimensional space to generate a consistent embedded state, which is then used for branch training in multi-class classification. The following provides further details:

### A. Implicit Branch

The goal of the Implicit Branch is to capture potential implicit representations in surveillance videos through the processing of visual features. This branch is divided into three parts: feature extraction of the scene channel, feature extraction of the behavior channel, and lite time-series module.

***Scene Channel*** The purpose of designing scene channels is to obtain the representation of each frame in the video. Let $V = \{f_1, f_2, \cdots, f_n\}$ denote the video sequence consisting of $n$ frames, where each frame $f_j \in \mathcal{F}$ for $j \in \{1,2,\cdots,n\}$. Each frame is processed using a frozen CLIP Image Encoder to extract frame-level features. Let $\mathcal{F}$ represent the space of frames and $\mathcal{F}_{clip}$ represent the feature space derived by the CLIP Image Encoder. Define $\Phi : \mathcal{F} \rightarrow \mathcal{F}_{clip}$ as the feature extraction function implemented by the CLIP Image Encoder. Therefore, the frame-level feature representations are given by $V_{clip} = \{\Phi(f_1), \Phi(f_2), \cdots, \Phi(f_n)\} = \{g_1^{scene}, g_2^{scene}, \cdots, g_n^{scene}\}$, where $g_j^{scene} = \Phi(g_j) \in \mathcal{F}_{clip}$ denotes the scene channel feature extracted from the $j$-th frame by frozen CLIP Image Encoder.

***Behavior Channel.*** The purpose of designing behavior channels is to capture changes between frames content to guide the model in identifying action changes within a scene. Previous studies typically utilized optical flow methods [34][35] to monitor action change. However, these methods have two main issues: first, optical flow may not be accurate in some violent scenes occurring at dusk. Second, the computational cost of optical flow is relatively high. To address this problem, an Efficient Behavior Monitoring Module (EBMM) is proposed to achieve fast and effective action change monitoring.

The proposed EBMM operates a two-stage monitoring mechanism. In the first stage, it captures motion changes by computing differences between consecutive frames. In the second stage, it removes background information by calculating changes between neighboring patches within the current frame.

Let $p_i^k$ denote the $k$-th patch in the $i$-th frame, for a subset of the feature maps $V_{clip} = \{g_1^{scene}, g_2^{scene}, \cdots, g_n^{scene}\}$. The proposed EBMM divides each frame into $m$ patches, resulting in a corresponding patch set $P_i = \{p_i^1, p_i^2, \ldots, p_i^m\}$. In the first stage, the proposed EBMM computes the motion difference for each patch by calculating the change between consecutive frames $\triangle p_i^k = p_i^k - p(i-1)^k$, where $i = 2, \cdots, n$ and $k = 1, \ldots, m$.

In the second stage, the proposed EBMM removes background effects. Let $N(k)$ denote the neighboring patch which consists of patches adjacent to $p_i^k$. For each neighboring patch $p_i^l$ where $l \in N(k)$, calculate the absolute difference in That is, $\delta p_i^{k,l} = \left| p_i^k - p_i^l \right|$. The background suppression value and aggregation is $\delta p_i^k = p_i^k - \frac{1}{|N(k)|} \sum_{l \in N(k)} p_i^l$, the total difference for each patch is then given in $S_i^k = \left| \triangle p_i^k \right| + \left| \delta p_i^k \right|$.

Let $A_i$ denote the weight scores computed by applying an exponential kernel function over the total differences $S_i^k$. The calculation of $A_i$ as shown in $A_i = e^{\alpha S_i^k} / \sum_j e^{\alpha S_i^k}$, herein, the $\alpha$ denotes the Hyperparameters. The weight scores are expanded to match the original patch structure. These expanded scores are then used to adjust the total difference scores $S_i^k$ for all patches in the frame. Let $S_i^{Behavior}$ denote the final representation, the calculated in $S_i^{Behavior} = A_i^\top \cdot S_i^k$. The $S_i^{Behavior}$ can receive higher weights, while background patches are weighed close to zero. Unlike $g_j^{scene}$, where all pixels in each frame have nearly equal influence in violent surveillance scenes, $S_i^{Behavior}$ places a strong emphasis on locations of violent behavior.

We analyze the computational complexity of the proposed Efficient Behavior Monitoring Module (EBMM) compared to optical flow methods [34], [35] in Theorem 1.

**Theorem 1**: *The computational complexity of EBMM is lower than that of optical flow methods [34] , [35], making it more lightweight.*

**Proof 1.** Optical flow methods [34], [35]: Optical flow methods compute motion vectors for each pixel in a frame of size $H \times W$. The complexity is $\mathcal{O}(n \cdot I \cdot HW \cdot K)$, where $n$ is the number of frames, $I$ is the number of iterations, $HW$ is the number of pixels, and $K$ is the kernel size (e.g., $K = 25$). The EBMM divides each frame into $m$ patches, with complexity $\mathcal{O}(n \cdot m \cdot d)$, where $m$ is the number of patches $m \ll HW$, and $d$ is the feature dimension per patch $d \ll HW$.

Since $m \cdot d \ll HW \cdot K \cdot I$, consider a 1080p video where $HW \approx 2 \times 10^6$, $m = 1.25 \times 10^5$, $d = 64$, $I = 10$, $K = 25$. Then, the EBMM complexity is $n \cdot 8 \times 10^6$, the optical flow complexity is $n \cdot 5 \times 10^8$. The ratio is $\frac{5 \times 10^8}{8 \times 10^6} = 0.016$. The computational complexity of EBMM is significantly lower than that of optical flow methods, proving its lightweight advantage.∎

Since the relationship between $S_i^{Behavior}$ and $g_j^{scene}$ is complementary, they are added together and fed into a lightweight temporal sequence module for further processing.

***Lite time-series module*** Till now, the proposed scene channel and behavior channel extract frame-level background and action features using the CLIP encoder. These features contain instantaneous information but lack the global temporal context crucial for WVAD tasks. In this section, a lightweight temporal sequence module is proposed, which is similar to the standard <u>R</u>eceptance <u>W</u>eighted <u>K</u>ey <u>V</u>alue (RWKV) [36] encoder, including self-attention, layer normalization (LN), and feed-forward network (FFN). Since the positions of frames in sequence data are already fixed, there is no need to apply positional encoding when training the temporal sequence module. The main difference between this approach and traditional RWKV or other Transformer-like encoders lies in the computation of self-attention, which is based on relative distance rather than feature similarity. This mechanism makes the model more lightweight, as it effectively reduces the complexity of attention computation by considering only the relative distance between frames, without the need to evaluate high-dimensional feature similarity. This helps reduce computational costs and makes training and inference more efficient.

Specifically, in the design of self-attention, let $\varkappa$ denote adjacency matrix. Since the similarity between $i$-th and $j$-th frames is only determined by their relative temporal distance, the value of $\varkappa$ is calculated as $\varkappa = (-|i-j|)/\sigma$, where $\sigma$ is a hyper-parameter to control the range of influence of distance relation. In this work, the input of *Lite time-series module* is the summation feature of $f_j^{scene}$ and $S_i^{Behavior}$.

The operation of the time series module can be treated as a two-stage process. In the first stage, the time series model utilizes the attention mechanism of $\varkappa$, computed through *Softmax* and *Layer Normalization* (LN), to incorporate temporal information into the features, thereby analyzing the input time series. Let $\Psi$ denote the feature with temporal information fused after the first step. The operation of the first stage is represented as shown in Exp. (1):

$$\Psi = \mathrm{LN}\left(Softmax\left(\varkappa\right)\left(S_i^{Behavior} + f_j^{scene}\right)\right), \qquad (1)$$

In the second stage, $\Psi$ is processed through a *Feed-Forward Network* (*FFN*), followed by a residual connection, and then through Layer Normalization to obtain the result. Let $\psi$ denote the output after the non-linear transformation and residual connection in the second step. The operation of the second stage is represented in Exp. (2):

$$\psi = \mathrm{LN}\left(FFN\left(\Psi\right) + \Psi\right). \qquad (2)$$

After completing the time series processing, in the Implicit Branch, a coarse-grained binary classifier is integrated. The classification objective function is adopted for the classification branch, which is presented in Exp. (3):

$$\mathcal{L}_{BCE} = -\frac{1}{N}\sum_{i=1}^{N}\left[y_i \log(\hat{y}_i) + (1-y_i)\log(1-\hat{y}_i)\right], \qquad (3)$$

where $\mathcal{L}_{BCE}$ represents the cross-entropy loss, $N$ is the number of samples, $y_i$ is the true label of the $i$-th sample and $\hat{y}_i$ is the predicted probability for the $i$-th sample.

In Theorem 2 and Theorem 3, we analyzed two key aspects: The computational complexity advantages of using index-based pairwise distance calculations and the lower computational complexity of RWKV compared to standard Transformers for sequence modeling. Here are the specific details:

**Theorem 2:** *The computational complexity of computing pairwise index-based distances is lower than computing pairwise feature-based similarities.*

**Proof 2.** Consider a sequence of length $m$, where each element corresponds to a frame with features in a $d$-dimensional space. To compute pairwise index-based distances, such as the relative temporal distance $|i-j|$ between frames $i$ and $j$, the operation requires evaluating the absolute difference of their indices. For each pair $(i, j)$, this computation takes $\mathcal{O}(1)$, operations, as it involves a single subtraction and absolute value calculation. With $m$ frames, there are $m^2$ pairs, leading to a total complexity of $\mathcal{O}(m^2) \cdot \mathcal{O}(1) = \mathcal{O}(m^2)$.

In contrast, computing pairwise feature-based similarities, as in standard self-attention mechanisms typically involves the dot product of feature vectors. For two frames $i$ and $j$, with feature vectors $x_i, x_j \in \mathbb{R}^d$, the dot product $x_i \cdot x_j$ requires $d$ multiplications and $d-1$ additions, totaling $\mathcal{O}(d)$ operations. For all $m^2$ pairs, the complexity becomes in $\mathcal{O}(m^2) \cdot \mathcal{O}(d) = \mathcal{O}(m^2 d)$.

Since the feature dimension $d > 1$, , it follows that $\mathcal{O}(m^2) < \mathcal{O}(m^2 d)$. Thus, the complexity of computing pairwise index-based distances is strictly lower than that of computing pairwise feature-based similarities. ∎

**Theorem 3:** *The computational complexity of using RWKV for sequence modeling is lower than that of using a standard Transformer.*

**Proof 3.** In a standard Transformer, the self-attention mechanism computes similarities between all pairs of elements in a sequence of length $m$, with each element represented by a $d$-dimensional feature vector. The process involves:

1. Projection: Computing query $Q = XW_q$, key

$K = XW_k$, and $V = XW_v$, where $X \in \mathbb{R}^{m \times d}$ and $W_q, W_k, W_v \in \mathbb{R}^{d \times d}$. Each matrix multiplication is $\mathcal{O}(md^2)$, totaling $\mathcal{O}(md^2)$ for all three.

2. Attention Scores: Computing $QK^T \in \mathbb{R}^{m \times m}$ requires $\mathcal{O}(m^2 d)$ operations.

3. Output: Applying softmax and computing $soft \max\left(QK^T / \sqrt{d}\right)V$ requires $\mathcal{O}(m^2)$ for softmax and $\mathcal{O}(m^2 d)$ for the matrix-vector multiplication.

The total complexity is:

$$\mathcal{O}(md^2) + \mathcal{O}(m^2 d) + \mathcal{O}(m^2 d) = \mathcal{O}(md^2 + m^2 d).$$

For large $m$ and $d$, this is dominated by $\mathcal{O}(m^2 d)$.

In contrast, RWKV employs an efficient attention mechanism, often formulated recurrently, avoiding the quadratic dependency on sequence length. In the given context, the attention is based on a predefined adjacency matrix $\varkappa$, where $\varkappa = (-|i-j|)/\sigma$, the operation $\Psi = \mathrm{LN}\left(Softmax(\varkappa)(S_i^{Behavior} + f_j^{scene})\right)$, involves:

1. Softmax on $\varkappa$: For $\varkappa \in \mathbb{R}^{d \times d}$, computing $Softmax(\varkappa)$ takes $\mathcal{O}(m^2)$ operations.

2. Matrix Multiplication: Multiplying $Softmax(\varkappa) \in \mathbb{R}^{d \times d}$ by the input $(S_i^{Behavior} + f_j^{scene})\mathbb{R}^{m \times d}$ requires $\mathcal{O}(m^2 d)$ operations.

However, RWKV's design leverages a recurrent formulation, processing the sequence step-by-step, reducing the complexity to $\mathcal{O}(md)$. per layer by maintaining a state that aggregates temporal information linearly. For a sequence of length $m$, the total complexity becomes $\mathcal{O}(md)$.

Since $\mathcal{O}(md) < \mathcal{O}(m^2 d)$ for $m > 1$, complexity of RWKV is lower than that of the standard Transformer. ∎

The above provides the complete content of the implicit branch; next, the explicit branch will be introduced.

*B. Explicit Branch*

The goal of the explicit branch is to perform fine-grained multi-class classification by utilizing contrastive learning between text and video. This branch is divided into two parts: rule generation based on YOLO-World and data mining, and contrastive learning between text and video.

**Rule Generation.** The purpose of rule generation is to create more interpretable video descriptions. Previous studies [17][18][31][32] typically used LLMs to generate descriptions for each frame, followed by temporal sequence models to obtain a textual embedding representation for the entire video. However, these methods have two main issues. First, the embedding representation is usually high-dimensional and lacks interpretability. Second, the LLMs tend to be unstable, making it difficult to ensure consistent scene descriptions for each frame. To address these problems, this paper employs a two-stage approach that combines object monitor and data mining to generate interpretable video descriptions. The following are the specific details:

Firstly, object detection is considered. In practice, the pre-trained model YOLO-World [20] is used to monitor the actions and objects presented in each frame. YOLO-World is a model pre-trained on a large-scale dataset through vision-language modeling, capable of efficient zero-shot detection. Specifically, given a video frame $I$ and a set of categories $U = \{U_1, U_2, \cdots, U_j\}$ that need to be monitored, the model can automatically identify and outline the corresponding objects in the video frame $I$. For a video $V = \{f_1, f_2, \cdots, f_n\}$, the model monitors each frame, resulting in $n$ pieces of data containing the objects monitored in each frame and their corresponding counts.

The goal of data mining here is to derive association rules from the given data. For instance, by using YOLO-World to analyze a video of a fight, it is possible to extract the ongoing actions and objects involved. By calculating the confidence and support of the objects appearing in each frame, the rule:

$$\{People\}\{Strick\} -> \{Fight\},$$

can ultimately be derived. This means that whenever people and sticks are present, it is likely that a fight is occurring. By referring the FP-Growth [37], this paper further improves efficiency in terms of the required time for data mining. Specifically, FP-Growth is a two-phase process involving the construction and pruning of an FP-Tree. Data mining is conducted through conditional judgments, tree structures, and frequency counting. However, using GPUs does not significantly enhance the efficiency of these operations, as they primarily rely on logical operations and conditional evaluations rather than parallel matrix computations. On the other hand, traditional CPUs experience efficiency degradation due to I/O constraints [38].

To address these issues, this paper proposes a multi-CPU parallel FP-Growth method. In this approach, data are divided into separate blocks and are processed in parallel across different CPU threads to construct multiple FP-Trees, which are eventually merged into a complete tree for data mining. The following defines the notations that will be used to present the proposed MCP-FP Algorithm. The detailed operations are given in the MCP-FP Algorithm.

*Definitions:*

| Notations | Meaning |
|---|---|
| $D$ | Dataset |
| $\{D_1, D_2, \ldots, D_n\}$ | Non-overlapping data blocks, where $D = \bigcup_{i=1}^{n} D_i$ and $D_i \cap D_j = \emptyset$ for $i \neq j$. |
| $min\_support \in [0,1]$ | Minimum support threshold. |
| $min\_confidence \in [0,1]$ | Minimum confidence threshold. |
| $freq_i(x)$ | Frequency (count) of item |
| $support(x) = \sum_{i=1}^{n} freq_i(x)$ | support of item $x$ in the entire dataset $D$. |
| $L_i$ | local frequent itemset for $D_i$, where |
| $FP\_Tree_i$ | $L_i = \{x|x \in D_i, freq_i(x) \ge min\_support\}$. FP-trees were constructed from $L_i$ and $D_i$. |
| $FP\_Tree_{Global}$ | Global FP-tree constructed by merging all $FP\_Tree_i$. |
| $F$ | Set of global frequent itemset $min\_support$ |
| $R$ | set of association rules satisfying $min\_support$ and $min\_confidence$. |

---

**MCP-FP Algorithm: Multi-CPU Parallel FP-Growth Construction**

**Input:** Dataset $D = \bigcup_{i=1}^{n} D_i$, divided into multiple chunks $\{D_1, D_2, \cdots, D_n\}$, min_support $\in [0,1]$, and min_confidence $\in [0,1]$.

**Output:** Set of association rules $R$.

Initialize $R \leftarrow \emptyset$ and $F \leftarrow \emptyset$
**For** $i = 1$ **to** $n$ **do:**
 Assign $D_i$ to CPU $i$.
**Parallel for** $i = 1$ **to** $n$ **do:**
 **For** each item $x \in D_i$:
  $freq_i(x) \leftarrow$ count of $x$ in $D_i$.
  $support(x) \leftarrow support(x) + freq_i(x)$
 $L_i = \{x|support(x) \ge min\_support \times |D|\}$
 $FP\_Tree_i \leftarrow$ ConstructFPtree$(L_i, D_i)$
Initialize $FP\_Tree_{Global} \leftarrow$ empty $FP\_Tree$
**For** $i = 1$ **to** $n$ **do:**
 **For** each node $x$ **in** $FP\_Tree_i$:
  **If** $x$ exists in $FP\_Tree_{Global}$:
   $FP\_Tree_{Global}(x) \leftarrow FP\_Tree_{Global}(x) + FP\_Tree_i(x)$
  **else**
   Add node $x$ to $FP\_Tree_{Global}$ with count $FP\_Tree_i(x)$
**Parallel for** each FP-tree $T_{conditional}$ derived from $FP\_Tree_{Global}$ **do:**
 $F \leftarrow F \cup$ MineFrequentItemsets $(T_{conditional}, min\_support)$
 **For** each frequent itemset $f \in F$:
  **For** each non-empty subset $s \subset f$:
   conf $\leftarrow \frac{support(f)}{support(s)}$
  **If** conf $\ge$ min_confidence:
   $R \leftarrow R \cup \{s \rightarrow f \setminus s\}$
**Output** $R$

---

***Contrastive Learning*** In weakly supervised video anomaly detection tasks, traditional single-modal information often has limitations in recognizing specific violent behaviors. Therefore, this paper explores multi-modal information to improve recognition performance. Cross-modal learning is a powerful approach because it can integrate data from different sources, such as visual and textual information, providing richer semantic understanding and stronger recognition capabilities.

This paper introduces the method of contrastive learning which aims to enhance the model's recognition ability by maintaining consistency across different modalities. Here, CLIP is a typical cross-modal contrastive learning model, which consists of two encoders: an image encoder and a text encoder. The main function of CLIP is to transform images and text into vectors and calculate their similarity, ensuring that the representations of related images and text are close to each other in the vector space. This enables the vectors of images and text to query each other, achieving full fusion of the vector representations in both image and text spaces, resulting in more comprehensive multi-modal recognition.

This paper uses both visual and textual information for recognition. One challenge is to effectively obtain enough textual information. Typically, cameras used for video recording do not support audio recording or audio-to-text conversion. The lack of sufficient textual descriptions can negatively impact the effectiveness of cross-modal learning. As a result, it is crucial to consider how to address this issue.

In the text branch, this paper designs two inputs. One input employs YOLO-World in combination with parallel FP-Growth to mine video descriptions, while the other is a learnable prompt used for dynamically adjusting the output to more accurately capture key features and content in the video. Let $T^{Rule} = \{T_1^{Rule}, T_2^{Rule}, \cdots, T_m^{Rule}\}$ denote the video descriptions. Let $T^{Prompt} = \{T_1^{Prompt}, T_2^{Prompt}, \cdots, T_m^{Prompt}\}$ denote the Learnable prompt. These two inputs are combined as the shared input for the CLIP text encoder, yielding $\phi^{text} = \{\phi_1^{text}, \phi_2^{text}, \cdots, \phi_m^{text}\}$. For image branch an embedding representation is directly adopted, using $\psi^{video} = \{\psi_1^{video}, \psi_2^{video}, \cdots, \psi_m^{video}\}$.

The alignment matrix $M_{m \times m} = [s_{i,j}]$ consists of the similarities between $m$ video features and $m$ text features, as shown in the following Exp. (4):

$$M_{m \times m} = \begin{array}{c} \\ F_1^{text} \\ \vdots \\ F_m^{text} \end{array} \overset{\displaystyle \psi_1^{video} \quad \cdots \quad \psi_m^{video}}{\begin{bmatrix} s_{1,1} & \cdots & s_{1,m} \\ \vdots & \ddots & \vdots \\ s_{m,1} & \cdots & s_{m,m} \end{bmatrix}}. \quad (4)$$

The similar value of $s_{i,j}$ in the $i$-th row and $j$-th column of $M$ can be calculated in $p_{i,j} = \exp(s_{i,j}/\tau)/\sum_j \exp(s_{i,j}/\tau)$, where $\tau$ is a scaling hyper parameter. The loss function of the entire model includes the alignment loss $\mathcal{L}_{align}$ and contrastive loss $\mathcal{L}_{contrast}$. The alignment loss $\mathcal{L}_{align}$ is shown in Exp. (5):

$$\mathcal{L}_{align} = -\sum_{i=1}^{m} y_i \log(p_i) + \lambda_1 \theta_{align}^2, \quad (5)$$

where $y_i$ represents the labels, $\lambda_1$ controls the regularization of the model parameters in the alignment loss, and $\theta_{align}$ represents the model parameters related to the alignment loss.

The contrastive loss $\mathcal{L}_{contrast}$ calculates the cosine similarity between normal class embeddings and other abnormal class embeddings, as shown in Exp. (6):

$$\mathcal{L}_{contrast} = \sum_j \max\left(0, \frac{t_n \cdot t_{aj}}{t_{n2} \cdot t_{aj2}} - \delta\right) + \lambda_2 \theta_{contrast}^2, \quad (6)$$

where $t_n$ represents the embeddings of the normal class, $t_{aj}$ represents the embeddings of the abnormal class, $\delta$ is a hyper parameter, $\lambda_2$ is a regularization parameter, and $\theta_{contrast}$ represents the model parameters related to the contrastive loss. The final total loss function is shown in Exp. (7):

$$\mathcal{L}_{total} = \mathcal{L}_{align} + \mathcal{L}_{contrast} + \mathcal{L}_{BCE}. \quad (7)$$

Through this design, the implicit and explicit branches are used for collaborative training and are integrated into an end-to-end system, ensuring that each loss function plays a unique role in the optimization process, aiming to prevent overfitting. During model training, the CLIP encoders are frozen, and the model fine-tunes both branches through backpropagation.

The above describes the entirety of the proposed RuleVAD system. In the next section, the experiments and model performance will be introduced.

## V. MODEL PERFORMANCE

In this section, relevant experimental performance and analysis are presented.

### A. Dataset and Evaluation Metrics

The experiments utilized two popular datasets from the WVAD domain: UCF-Crime [6] and XD-Violence [21]. XD-Violence, the largest publicly accessible dataset in violence monitoring, contains 4,754 videos amounting to 217 hours and spans six categories of violent incidents: verbal abuse, car accidents, explosions, fights, riots, and shootings. This dataset is randomly split into a training set with 3,954 videos and a test set with 800 videos, where the test set includes 500 violent and 300 non-violent videos. The UCF-Crime dataset comprises 1,900 real-world surveillance videos, with 1,610 videos designated for training and 290 for testing.

In terms of evaluation metrics, consistent with previous studies [4][6][8] [17] [18] [22] [23] [32], in XD-Violence dataset, we use the $\mathcal{AP}^{\mathcal{M}}$ evaluation the model performance. The Exp. (8) shows the $\mathcal{AP}^{\mathcal{M}}$ calculation.

$$\mathcal{AP}^{\mathcal{M}} = \sum_{k=1}^{K-1} \left( \mathcal{R}^{\mathcal{M}}\left(\theta_{k+1}\right) - \mathcal{R}^{\mathcal{M}}(\theta_k)\right) \times P\left(\theta_{k+1}\right)^{\mathcal{M}}, \quad (8)$$

this metric is computed by summing over a series of $K-1$ discrete confidence threshold points, indexed by $k$. For each step in the summation, the increase in recall, represented by the term $\mathcal{R}^{\mathcal{M}}\left(\theta_{k+1}\right) - \mathcal{R}^{\mathcal{M}}(\theta_k)$, is multiplied by the corresponding precision value, $P\left(\theta_{k+1}\right)^{\mathcal{M}}$, which is measured at the next threshold level $\theta_{k+1}$. In this formulation, $\mathcal{R}^{\mathcal{M}}$ and $P^{\mathcal{M}}$ signify the recall and precision for the model or class $\mathcal{M}$ at a given threshold. This method effectively calculates the area under the Precision-Recall curve by aggregating the precision values at each incremental change in recall.

In UCF-Crime dataset, we use the $\mathcal{A}_{\mathcal{M}}$ evaluation the model performance. The Exp. (9) shows the $\mathcal{A}_{\mathcal{M}}$ calculation.

$$\mathcal{A}_{\mathcal{M}} = \int_0^1 \frac{TP(\theta)}{TP(\theta)+FN(\theta)} d\left(\frac{FP(\theta)}{FP(\theta)+TN(\theta)}\right), \quad (9)$$

herein, $TP(\theta)/TP(\theta)+FN(\theta)$ is the True Positive Rate (TPR), It measures the proportion of actual positive samples that are correctly identified by the model. $FP(\theta)/FP(\theta)+TN(\theta)$ is the False Positive Rate (FPR). It measures the proportion of actual negative samples that are incorrectly classified as positive.

### B. Implementation Details

In practice, this study follows previous studies [17][18][32] by using a frozen CLIP (ViT-B/16) to extract features from video frames. For the XD-Violence [6] and UCF-Crime datasets [21], the experiment processes 1 out of 16 frames on UCF-Crime dataset. During the training stage, the maximum length of training videos is set to 256; videos exceeding this limit are sampled down to this length. For all datasets, the length of the learnable prompts prefixed to text labels is set to 8. For each image patch $p_i$, the image is resized to $224 \times 224$, and a sliding window of $32 \times 32$ with a stride of 32 is used to generate multiple patches.

For hyperparameter settings, $\alpha$ is uniformly set to 0.2. The $\lambda_1$ is $5 \times 10^{-4}$ for the XD-Violence dataset and $1 \times 10^{-3}$ for the UCF-Crime dataset. The $\lambda_2$ is $6 \times 10^{-4}$ for the XD-Violence dataset and $2 \times 10^{-3}$ for the UCF-Crime dataset. The $\sigma$ in adjacency matrix $\varkappa$ is 0.8. For model optimization, the AdamW optimizer with a learning rate of 1e-4.

### C. Comparison with State-of-the-art Methods

The proposed RuleVAD can simultaneously realize coarse-grained and fine-grained violence monitoring. The experiment presents the performance of the proposed RuleVAD and compares it with several state-of-the-art methods on coarse-grained and fine-grained WVAD tasks.

TABLE I

COARSE-GRAINED COMPARISONS ON XD-VIOLENCE ( $\mathcal{AP}^{\mathcal{M}}$ ) AND UCF-CRIME ( $\mathcal{A}_{\mathcal{M}}$ ).

| Category | Method | $\mathcal{AP}^{\mathcal{M}}$ | $\mathcal{A}_{\mathcal{M}}$ |
|---|---|---|---|
| Semi supervised | SVM baseline | 50.80 | 50.10 |
| | OCSVM [23] | 28.63 | 63.20 |
| | C3D [22] | 31.25 | 51.20 |
| Weak supervised | CLIP [8] | 76.57 | 84.72 |
| | I3D [6] | 75.18 | 84.14 |
| | HL-Net [4] | 80.00 | 84.57 |
| | CLIP-TSA [32] | 82.17 | 87.58 |
| | VadCLIP [18] | 84.51 | 88.02 |
| | STPrompt [17] | 85.22 | 88.08 |
| | **RuleVAD (Ours)** | **87.48** | **89.63** |

Table I compares the performance of the proposed RuleVAD with other approaches on the XD-Violence and UCF-Crime datasets for coarse-grained monitoring. The results show that the proposed RuleVAD performs excellently on both datasets, significantly outperforming other semi-supervised and classification-based weakly supervised methods. Overall, RuleVAD achieves 87.48% $\mathcal{AP}^{\mathcal{M}}$ , on the XD-Violence dataset and 88.58% $\mathcal{A}_{\mathcal{M}}$ on the UCF-Crime dataset, reaching the current state-of-the-art results. Compared to its strongest

TABLE II

Fine-grained comparisons on XD-Violence and UCF-Crime.

| Method | Mean $\mathcal{AP}^{\mathcal{M}}$ @IoU (%) (XD-Violence.) | | | | | | Mean $\mathcal{A}_{\mathcal{M}}$ @IoU (%) (UCF-Crime.) | | | | | |
|---|---|---|---|---|---|---|---|---|---|---|---|---|
| | 0.1 | 0.2 | 0.3 | 0.4 | 0.5 | AVG | 0.1 | 0.2 | 0.3 | 0.4 | 0.5 | AVG |
| Random Baseline | 1.82 | 0.92 | 0.48 | 0.23 | 0.09 | 0.71 | 0.21 | 0.14 | 0.04 | 0.02 | 0.01 | 0.08 |
| SVM | 18.64 | 12.53 | 11.05 | 8.26 | 4.32 | 10.96 | 3.64 | 3.59 | 2.39 | 1.34 | 0.98 | 2.39 |
| OCSVM [23] | 19.85 | 9.06 | 8.56 | 6.55 | 6.03 | 10.01 | 4.85 | 4.44 | 2.33 | 1.85 | 1.56 | 3.01 |
| C3D [22] | 20.28 | 13.72 | 8.44 | 5.06 | 2.81 | 10.06 | 4.33 | 3.88 | 2.46 | 1.66 | 2.11 | 2.89 |
| I3D [6] | 22.72 | 15.57 | 9.98 | 6.20 | 6.78 | 12.25 | 5.73 | 4.41 | 2.69 | 1.93 | 1.44 | 3.24 |
| HL-Net [4] | 35.35 | 28.02 | 20.94 | 15.01 | 10.33 | 21.93 | 8.47 | 6.88 | 5.62 | 4.91 | 3.38 | 5.85 |
| CLIP-TSA [32] | 34.53 | 32.88 | 28.11 | 13.65 | 10.01 | 23.84 | 12.62 | 8.13 | 6.66 | 4.28 | 1.91 | 6.72 |
| VadCLIP [18] | 37.03 | 30.84 | 23.38 | 17.90 | 14.31 | 24.70 | 11.72 | 7.83 | 6.40 | 4.53 | 2.93 | 6.68 |
| STPrompt [17] | 39.58 | 31.26 | **28.35** | 19.28 | **16.29** | 26.95 | 12.59 | **8.13** | 6.67 | **5.30** | **3.54** | 7.25 |
| **RuleVAD (Ours)** | **41.29** | **38.88** | 27.64 | **21.33** | 13.79 | **28.59** | **18.82** | 7.66 | **7.16** | 5.02 | 3.51 | **8.43** |

competitors, VadCLIP and STPrompt, the proposed RuleVAD improves the $\mathcal{AP}^{\mathcal{M}}$ metric on the XD-Violence dataset by 2.97% and 2.26%, respectively, and exceeds the $\mathcal{A}_{\mathcal{M}}$ metric on the UCF-Crime dataset by 1.61% and 1.55%, respectively.

This performance improvement is primarily attributed to the use of an implicit branch and multi-channel training strategy in the proposed RuleVAD. This approach jointly extracts frame-level spatial information through behavior and scene channels. Then the proposed RuleVAD employs a lightweight temporal module to analyze temporal relationships. The advantages of this design are that RuleVAD effectively separates and processes behavior and scene feature, avoids information conflicts, and significantly enhances model interpretability. Additionally, the end-to-end training approach allows coarse-grained tasks to leverage the knowledge from fine-grained implicit branches, resulting in more pronounced effects in coarse-grained monitoring tasks. Regarding the performance of fine-grained WVAD tasks, Table III presents a comparison of the RuleVAD system developed in this paper with other related studies, such as VadCLIP and STPrompt, on the XD-Violence and UCF-Crime datasets. For fairness, all feature extractors utilize CLIP (ViT-B/16). Compared to coarse-grained experiments, fine-grained tasks in WVAD are more challenging, as they require not only monitoring the occurrence of violent events but also accurately identifying the specific categories of violent types.

As shown in Table II, the proposed RuleVAD system outperforms other methods on the XD-Violence and UCF-Crime datasets. This advantage is primarily due to the input strategy employed in the RuleVAD system's text encoder. Unlike methods that directly use text generated by LLM, the proposed RuleVAD system combines association rules generated through data mining with learnable prompts as input. This approach not only produces a highly condensed and interpretable video description but also enhances the system's ability to identify violent events. Through this input strategy, the system focuses on task-relevant semantic space and captures subtle distinctions in events, resulting in greater accuracy and stability in fine-grained classification tasks.

*C. Ablation Studies*

Table III presents the ablation study on the three modules within the two branches of the proposed RuleVAD system. As shown in Table III, all three modules are essential across both benchmark datasets, XD-Violence and UCF-Crime. Compared with the related studies, when a single module is used, system performance improves, though the gains are limited compared to combining multiple modules. Notably, when the MCP-FP module is combined with either the EBMM or RWKV module, significant performance enhancements are observed, indicating that MCP-FP provides crucial information gain by mining association rules. When all three modules are applied simultaneously, the system achieves optimal performance, with high accuracy rates of 87.48 $\mathcal{AP}^{\mathcal{M}}$ value on XD-Violence and 89.63 $\mathcal{A}_{\mathcal{M}}$ value on UCF-Crime. These results show the importance of multi-module synergy for WVAD tasks.

TABLE III

Effectiveness of each module

| Components | | | XD-Violence | UCF-Crime |
|---|---|---|---|---|
| EBMM | RWKV | MCP-FP | $\mathcal{AP}^{\mathcal{M}}$ | $\mathcal{A}_{\mathcal{M}}$ |
| √ | | | 62.11 | 60.88 |
| | √ | | 63.34 | 58.65 |
| | | √ | 68.85 | 42.33 |
| √ | √ | | 72.44 | 69.88 |
| √ | | √ | 76.85 | 74.26 |
| | √ | √ | 73.59 | 72.11 |
| √ | √ | √ | **87.48** | **89.63** |

Table IV presents the ablation study on the designed EBMM module. As shown in Table IV, the dual-channel design achieves significant performance improvements on both benchmark datasets. When using only the scene or behavior channel individually, system performance reaches 81.35 and 84.61 $\mathcal{AP}^{\mathcal{M}}$ value on XD-Violence and 83.65 and 85.53 $\mathcal{A}_{\mathcal{M}}$ value on UCF-Crime, respectively. However, when both the scene and behavior channels are activated simultaneously, the system achieves optimal performance. These results indicate that the scene and behavior channels capture complementary aspects of abnormal features, and their combination enables a more comprehensive improvement in WVAD. Furthermore, these findings collectively validate the effectiveness of **Theorem 1**.

TABLE IV

ABLATION STUDIES ON EBMM MODULE

| Components | | XD-VIOLENCE | UCF-CRIME |
|---|---|---|---|
| Scene Channel | Behavior Channel | $\mathcal{AP}^{\mathcal{M}}$ | $\mathcal{A}_{\mathcal{M}}$ |
| √ | | 81.35 | 83.65 |
| | √ | 84.61 | 85.53 |
| √ | √ | **87.48** | **89.63** |

TABLE V

ABLATION STUDIES ON TIME SERIES MODULE

| Method | XD-VIOLENCE | UCF-CRIME |
|---|---|---|
| | $\mathcal{AP}^{\mathcal{M}}$ | $\mathcal{A}_{\mathcal{M}}$ |
| Transformer | 81.56 | 87.25 |
| Distance Transformer | 84.24 | 88.01 |
| RWKV | 85.33 | 88.34 |
| Distance RWKV | **87.48** | **89.63** |

Table V presents the ablation study on the time series module. As shown in Table V, different time series methods exhibit significant performance differences across the two benchmark datasets. Using the Transformer model, the system achieves 81.56 $\mathcal{AP}^{\mathcal{M}}$ value and 87.25 $\mathcal{A}_{\mathcal{M}}$ value on XD-Violence and UCF-Crime, respectively. When incorporating the distance-aware mechanism, the performance of the Distance Transformer improves, reaching 84.24 $\mathcal{AP}^{\mathcal{M}}$ value and 88.01 $\mathcal{A}_{\mathcal{M}}$ value on the two datasets. In comparison, the RWKV method further enhances performance, achieving 85.33 $\mathcal{AP}^{\mathcal{M}}$ value on XD-Violence and 88.34 $\mathcal{A}_{\mathcal{M}}$ value on UCF-Crime. The combination of RWKV with the distance-aware mechanism yields the best results. These results indicate that the distance-aware mechanism effectively enhances the capability of time-series models to detect anomalous events, while RWKV demonstrates significant advantages in capturing temporal dependencies. Furthermore, these findings collectively validate the effectiveness of **Theorem 2** and **Theorem 3**.

TABLE VI

COMPUTATIONAL COMPLEXITY COMPARISON

| Methods | Feature | Parameter | FLOPS |
|---|---|---|---|
| I3D [6] | I3D | 28M | 186.9G |
| C3D [22] | C3D | 78M | 386.2G |
| VadCLIP[18] | Clip | 36.5M | 82.3G |
| STPrompt [17] | Clip | 31.5M | 44.8G |
| **RuleVAD (Ours)** | Clip | 28.8M | 36.5G |

Table VI aims to present the parameter count (Parameter) and computational complexity (FLOPS) of the proposed RuleVAD, C3D, I3D, VadCLIP, and STPrompt on the XD-Violence dataset. The parameter count represents the number of weights in the model, which is used to measure the model's complexity and storage requirements, while FLOPS indicates the computational load required for a single forward pass. As shown in Table VI, the proposed RuleVAD uses a distance-based attention mechanism combined with an RWKV module, replacing the traditional embedding similarity calculation in Transformers. This lightweight design significantly reduces both the model's parameter counts and the computational cost for each forward pass.

It is noteworthy that current edge computing devices, such as the NVIDIA Jetson Nano, impose restrictions on model parameters and computational load. More specifically, the model parameters are limited to 100M, and the computational load is capped at 100G FLOPS, while the Google Coral Edge TPU restricts model parameters to 32MB. The proposed RuleVAD meets the operational requirements of both devices, making it more suitable for edge computing environments compared to other methods.

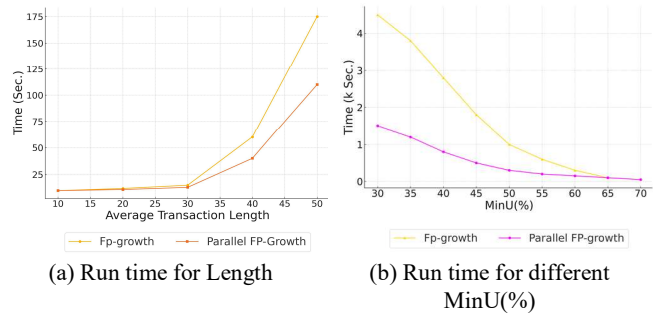

(a) Run time for Length  (b) Run time for different MinU(%)

Fig.4. Ablation studies on MCP-FP.

Figure 4 aims to present the ablation study of MCP-FP Algorithm, consisting of two subplots that validate the feasibility of the proposed method from two perspectives. Subfigure (a) displays the execution times of the Fp-growth and Parallel FP-Growth methods across different transaction lengths. The x-axis represents the average transaction length, ranging from 10 to 50, while the y-axis represents the execution time in seconds (Sec.). As the average transaction length increases, both methods show an upward trend in execution time. However, the growth rate of Parallel FP-Growth is notably slower than that of FP-Growth, indicating its higher efficiency.

Subfigure (b) shows the execution time comparison between FP-Growth and MCP-FP Algorithm under varying MinU(%) conditions. The x-axis represents the minimum usage threshold (MinU%) from 30% to 70%, and the y-axis represents the execution time in seconds. The graph illustrates that the execution time of both methods decreases with the MinU(%). However, Parallel FP-Growth consistently outperforms FP-growth especially at lower MinU(%) values. These results indicate that Parallel FP-Growth is highly efficient, particularly at lower support thresholds, where it significantly reduces time.

Table VII presents ablation studies on prompts in the XD-Violence dataset. This experiment compares the proposed RuleVAD with handcrafted prompts [33], learnable prompts [32], frame-level average prompts [18], and anomaly-specific prompts [17]. The results show that RuleVAD achieves the best performance. The primary reason for RuleVAD superior performance is that it leverages interpretable association rules mined directly from Yolo-World [20], combined with learnable parameters. This approach contrasts with other methods that rely on manually crafted labels or prompts generated by LLM, which may face issues such as limited interpretability, inconsistency, or overfitting to specific types of violence.

TABLE VII

ABLATION STUDIES ON PROMPT

| Prompt | AVG (%) |
|---|---|
| Hand-crafted Prompt | 18.46 |
| Learnable-Prompt | 23.84 |
| Average-Frame Visual Prompt | 24.70 |
| Anomaly-Focus Visual Prompt | 26.95 |
| **RuleVAD (Ours)** | **28.59** |

Figure 5 presents four examples, showing that when the crowd size increases, the level of violence escalates from "fighting" to "rioting." To validate this finding, this study conducted a statistical analysis of videos in the "fights" and "riots" categories within the XD-Violence dataset. Specifically, the experiment used Yolo-World to monitor and count each frame in the videos and then output the top 10 most important objects. Figure 6 shows the proportion of each object in two different scenes. Fig. 6 (a) represents the "Fight" scenario, where "Fist" has the highest proportion at 25%. Fig. 6 (b) represents the "riot" scenario, where the two most prominent elements in the scene are "People" and "Person". This observation suggests that as crowd size increases during violent incidents, it is advisable to trigger quick alerts to prevent escalation.

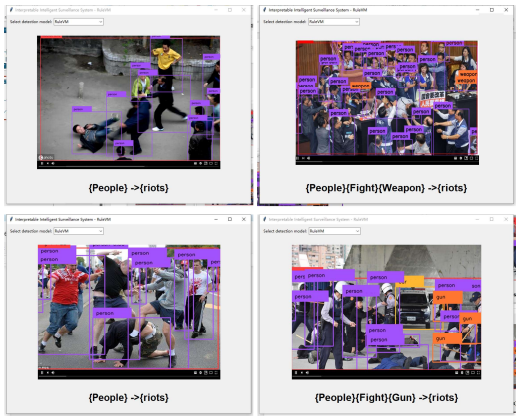

**Fig.5.** The proposed RuleVAD system visualization.

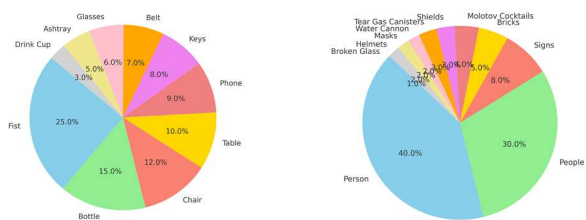

(a). Fight  (b). Riot

**Fig.6.** Top 10 important objects in Fight and Riot scene.

Table VIII presents a comparison of the performance of the proposed EBMM module under different hyperparameter $\alpha$ settings. As shown in Table IX, the best performance on the XD-Violence and UCF-Crime datasets is achieved when the parameter $\alpha$ is set to 0.2.

Figure 7 illustrates detection examples of the proposed RuleVAD across multiple video clips in XD-Violence. Each subplot displays the anomaly detection curve for a video segment, with red regions indicating detected anomalous behaviors. The figure includes videos from diverse scenes, such as a violent scene from "Deadpool 2 (2018)" and an explosion scene from "In the Head (1990)". Using the proposed RuleVAD, the system accurately identifies anomalous behaviors in the videos, such as violence (v-sbEHU4Gsbx4),

TABLE VIII

PERFORMANCE COMPARISON ANALYSIS OF THE EBMM MODULE UNDER DIFFERENT HYPERPARAMETER $\alpha$ SETTINGS.

| | XD-VIOLENCE $\mathcal{AP}^{\mathcal{M}}$ | UCF-CRIME $\mathcal{A}_{\mathcal{M}}$ |
|---|---|---|
| 0.1 | 86.50 | 88.70 |
| 0.2 | **87.48** | **89.63** |
| 0.3 | 87.01 | 89.21 |
| 0.4 | 86.20 | 88.50 |
| 0.5 | 85.33 | 87.80 |
| 0.6 | 84.57 | 87.06 |
| 0.7 | 83.71 | 86.24 |
| 0.8 | 82.92 | 85.52 |
| 0.9 | 82.13 | 84.81 |

explosions (v-BEPlatuSA), and shootings (v-3wMhrZ0rc), while marking the time intervals of anomalies. These results demonstrate that the proposed RuleVAD can effectively detect anomalous behaviors in videos, providing reliable anomaly recognition capabilities for surveillance systems.

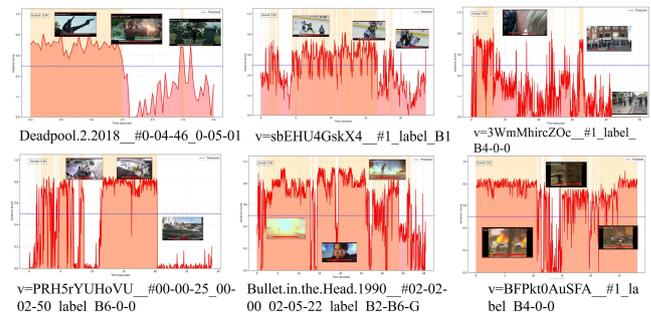

Deadpool.2.2018__#0-04-46_0-05-01  v-sbEHU4GskX4__#1_label_B1  v-3WmMhrZOc__#1_label_B4-0-0

v-PRH5rYUHoVU__#00-00-25_00-02-50_label_B6-0-0  Bullet.in.the.Head.1990__#02-02-00_02-05-22_label_B2-B6-G  v-BFPkt0AuSFA__#1_label_B4-0-0

**Fig.7.** Examples of the proposed RuleVAD.

## VI . CONCLUSION

In this paper, we introduced RuleVAD, a novel system designed to address the critical challenge of deploying high-performance, weakly-supervised video anomaly detection on resource-constrained consumer electronics. Our innovative decoupled dual-branch architecture effectively balances computational efficiency with analytical depth. The lightweight implicit branch rapidly filters normal events using only visual cues, conserving resources, while the multimodal explicit branch performs nuanced, fine-grained analysis on complex scenes when necessary. By combining YOLO-World object detection with data mining to generate interpretable rules, RuleVAD significantly enhances accuracy and reduces the false alarms that are a major concern for end-users.

Experimental results on the XD-Violence and UCF-Crime datasets confirm that RuleVAD substantially outperforms state-of-the-art methods. More importantly, its ability to identify meaningful patterns, such as the correlation between crowd size and violence, showcases its capacity to provide actionable insights. The successful deployment and real-time execution of the entire system on a low-power NVIDIA Jetson Nano board is a key contribution, proving its practical readiness for integration into consumer products like smart cameras, video doorbells, and personal safety devices.